\documentclass[10pt,twocolumn,letterpaper]{article}

\usepackage{iccv}
\usepackage{times}
\usepackage{epsfig}
\usepackage{graphicx}
\usepackage{amsmath}
\usepackage{amssymb}

\usepackage{tabularx}
\usepackage{color}
\usepackage{subfig}
\usepackage{booktabs}
\usepackage{amsthm}
\usepackage[T1]{fontenc}
\usepackage{multirow}
\usepackage{textpos}

\newcommand{\copyrightstatement}{
    \begin{textblock}{13}(0,9.5)    
        \centering
         \noindent
         \footnotesize
         \copyright 2021 IEEE. Personal use of this material is permitted. 
         Permission from IEEE must be obtained for all other uses, in any current or future 
  media, including \\reprinting/republishing this material for advertising or promotional 
  purposes, creating new collective works, for resale or redistribution to servers or\\ 
  lists, or reuse of any copyrighted component of this work in other works. 
    \end{textblock}
}

\usepackage[breaklinks=true,bookmarks=false]{hyperref}

\iccvfinalcopy 

\def\iccvPaperID{****} 

\ificcvfinal\fi

\begin{document}

\title{ Field-Based Plot Extraction Using UAV RGB Images}

\author{Changye Yang$^{1}$ \and Sriram Baireddy$^{1}$ \and Enyu Cai$^{1}$  \and Melba Crawford$^{2}$  \and Edward J. Delp$^{1}$  \\
$^{1}$Video and Image Processing Laboratory (VIPER), School of Electrical and Computer Engineering\\
$^{2}$School of Civil Engineering\\
Purdue University\\
West Lafayette, Indiana, USA


}



\maketitle
\copyrightstatement
\ificcvfinal\thispagestyle{empty}\fi


\begin{abstract}
Unmanned Aerial Vehicles (UAVs) have become popular for use in plant phenotyping of field based crops, such as maize and sorghum, due to their ability to acquire high resolution data over field trials.
Field experiments, which may comprise thousands of plants, are planted according to experimental designs to evaluate varieties or management practices.
For many types of phenotyping analysis, we examine smaller groups of plants known as ``plots.''
In this paper, we propose a new plot extraction method that will segment a UAV image into plots. 
We will demonstrate that our method achieves higher plot extraction accuracy than existing approaches.
\end{abstract}


\section{Introduction}
Plant phenotyping refers to the characterization and quantification of physical traits of plants such as height, leaf area, biomass, or flowering time~\cite{li_2014}.
Traditional phenotyping methods involve labor intensive field work~\cite{costa2019}. 
Modern high-throughput methods such as the use of Unmanned Aerial Vehicle (UAV) imaging~\cite{xie2020} can drastically reduce the workload and cost~\cite{yang2020high}. 
Recent technological improvements in imaging sensors, UAV platforms, and computational hardware also makes UAV systems more accessible~\cite{araus2013}. 
UAV-based imaging systems have been shown to work well for plant phenotyping tasks such as biomass prediction~\cite{tang2021}, salinity stress analysis~\cite{johansen2019}, and disease detection~\cite{wiesnerhanks2019}.

\begin{figure}[t]
	\centering
	\centerline{\includegraphics[width = 0.4\textwidth]{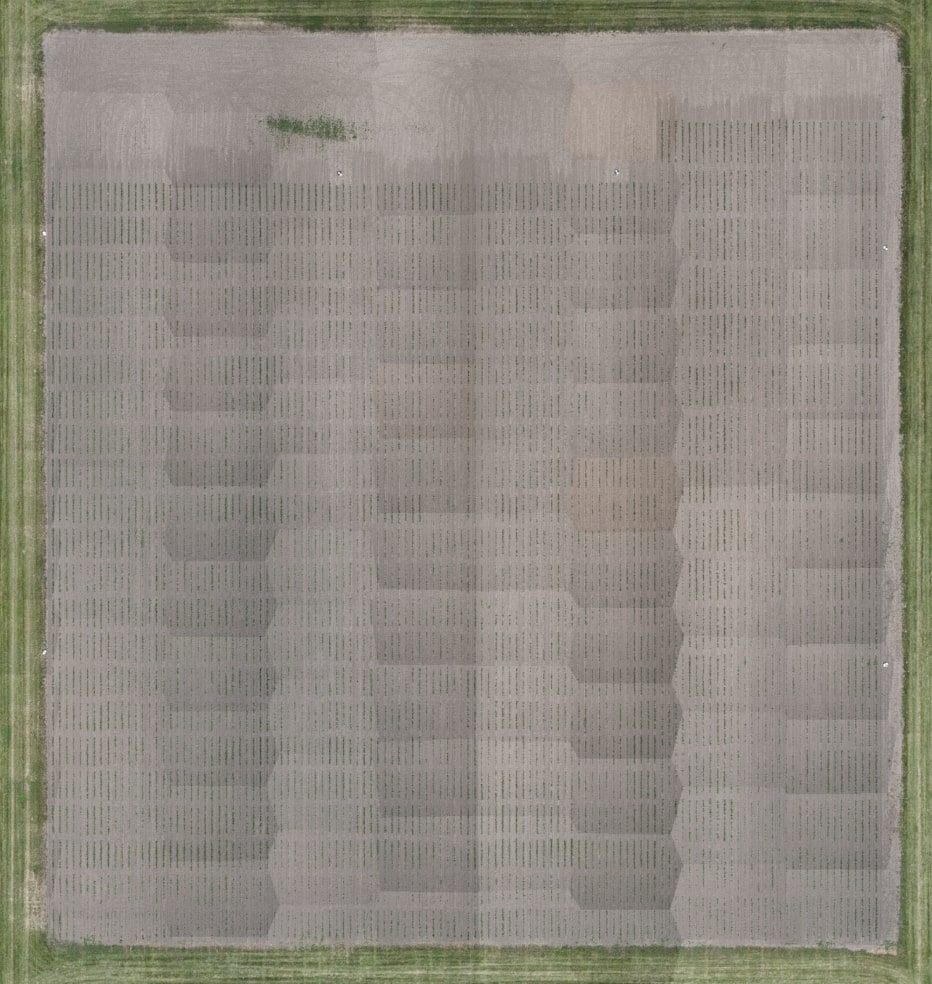}}
	\caption{RGB orthomosaic image acquired over sorghum with a resolution of 1cm/pixel}
\label{field_overview}
\end{figure}

\begin{figure}[t]
	\centering
	\centerline{\includegraphics[width = 0.4\textwidth]{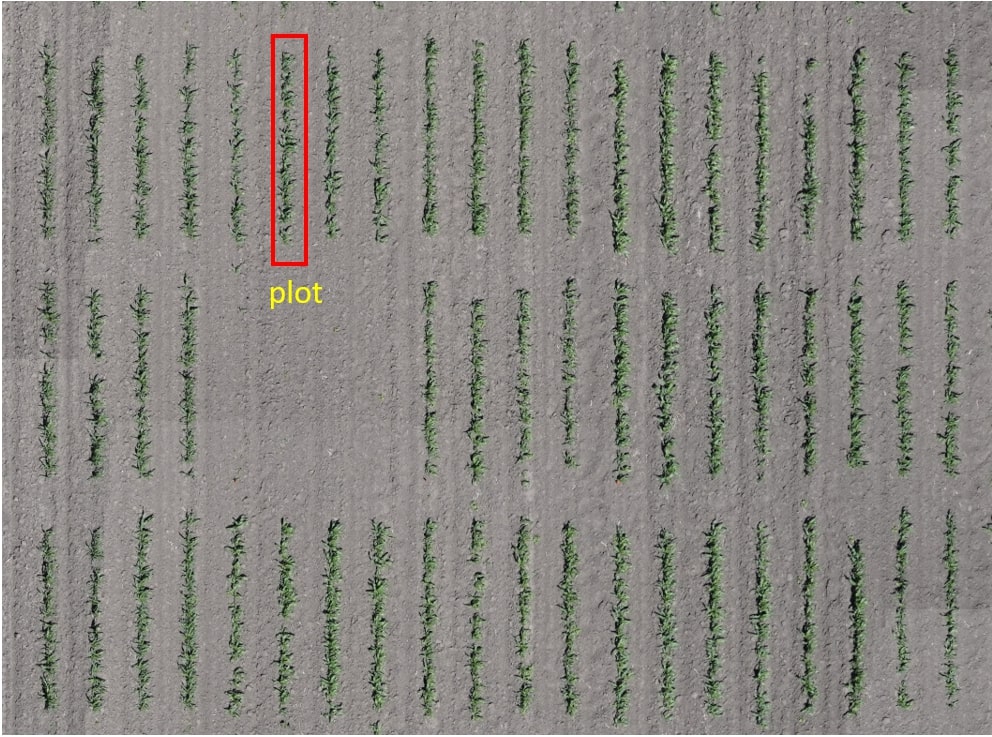}}
	\caption{Enlarged 1cm resolution orthomosaic image with plots visible (note the extracted plot shown in red)}
\label{closer_ortho}
\end{figure}

Images acquired from UAVs are typically processed to produce geometrically corrected, georeferenced orthomosaic images over an extended area~\cite{habib2016,lin2021}.
Figure~\ref{field_overview} shows an RGB orthomosiac image of a field of sorghum~\cite{vanderlip_1972} acquired at an altitude of 40 meters with spatial resolution of 1 centimeter/pixel.
A planted field consists of many smaller groups of plants  known as ``plots'' as shown in Figure~\ref{closer_ortho}.
Field trials for plant breeding experiments may comprise hundreds to thousands of plots within a field.  
Plot scale evaluation of phenotypes requires analysis of the responses within individual plots, which must be extracted.
Manual plot extraction is not feasible due to the very large number of plots in a field.
Using GPS-guided precision planters, fields are often planted in a ``grid'' pattern as shown in
Figure~\ref{closer_ortho}. 
These planting patterns will aid in plant extraction.

In this paper, we propose a plot extraction method to extract the plots from RGB UAV orthomosaic images.
We refer to our method as Comb Function Optimization Plot Extraction (COPE).
Our method utilizes a series of ``comb'' functions to locate the gaps between the plots by optimizing 2 energy functions related to the quantity of plants within a given area.
COPE can be used on almost any grid-like planted field (e.g., GPS-guided precision planter) and requires no pre-training.

\section{Background}

Several plot extraction methods have been described in the literature.  
In~\cite{tresch2019}, Tresch~\etal proposed a plot extraction method known as EasyMPE, 
which converts the field plant pixels into an energy map. 
Each plot is identified by thresholding the energy map. 
In theory, each plot will have two cutoff lines, one before the plot begins and one after the
plot ends. 
In practice, there might be not be exactly two cutoff lines due to variations in the energy function. 
This can result in missing or extra plots.

In~\cite{khan2019}, Khan~\etal proposed a plot extraction method that starts with a uniform-sized grid of plots. 
The position of each plot is then adjusted by minimizing the inter-plot energy and maximizing intra-plot energy. 
This method assumes the plots to be fixed size, which is limiting because this is not true for many fields.
In addition, this method fails if there are no plants visible in the plot: for example in Figure~\ref{closer_ortho}.
Possible causes for a lack of plants can be rain damage or no/late germination.

In \cite{javithesis}, Prat proposed a method assuming equal distance between plots. 
Prat first finds the binary plant segmentation mask from the orthomosaic image by thresholding the H color channel of the HSV color space~\cite{sharma2003}.
The resulting plant segmentation mask is then used to form energy functions that are used to find gaps in the plots.
Assuming an equal distance between plots, the boundaries of the plots are then found by optimization of the energy functions. 
The equal distance assumption for the plots causes the resulting extracted plot boundaries to not always lie in the gaps between plots; this approach then needs time-consuming manual adjustments to be useful.
Our proposed method resolves the issues described for the previous methods.

\begin{figure}[]
	\centering
	\centerline{\includegraphics[width = 0.4\textwidth]{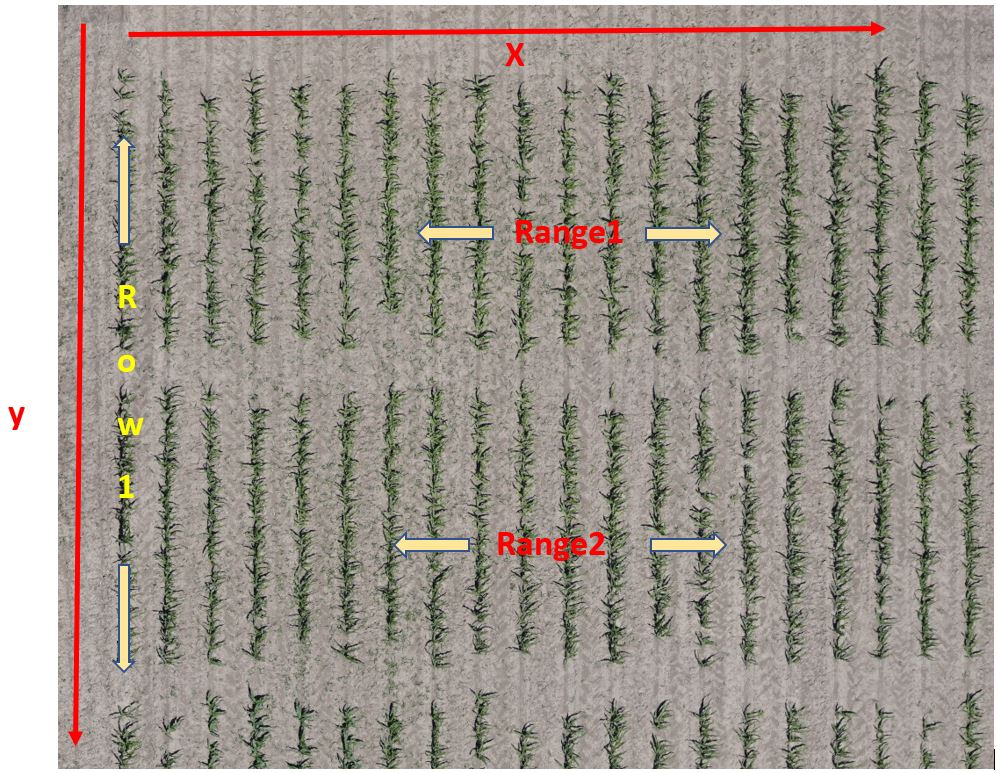}}
	\caption{The row and ranges shown for an enlarged orthomosaic image.}
\label{row_range}
\end{figure}

    


\section{Proposed Method: COPE}
We assume that the field has been planted in a grid pattern described above. 
For convenience, we further assume the field is planted in a north/south orientation as shown in Figure~\ref{row_range}.
We define a ``row'' in the orthomosaic image as a north/south (vertical) near-linear group of plots.
The ``range'' is defined to be  a east/west (horizontal) near-linear group of plots.
The planted field can be divided into rows and ranges as shown in Figure~\ref{row_range}.
Note a full row in the field will be associated with many plots, and similarly, a range will cross many rows.

For our work, a precision planter drops seeds for a given number of rows (denoted as $C$) simultaneously.  
The planter drives the length of the field, and the seeds of a given variety are dropped for the predetermined distance for a plot; no seeds are dropped between the ranges (creating gaps). 
The planter turns around at the end of the field and plants the next set of $C$ rows.





\begin{figure}[t]
	\centering
	\centerline{\includegraphics[width = 0.4\textwidth]{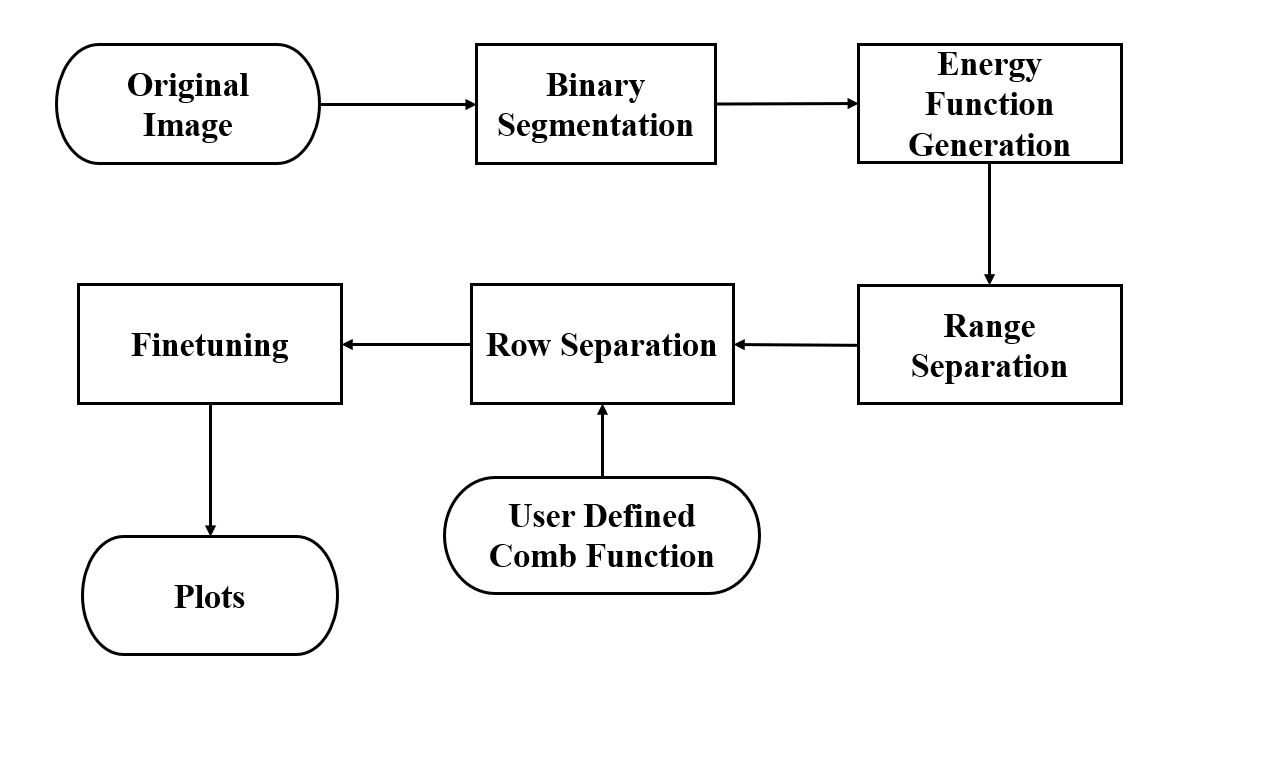}}
	\caption{Block diagram of COPE}
\label{block_diagram}
\end{figure}

Figure~\ref{block_diagram} shows a block diagram of COPE. 
The plants are segmented from the orthomosaic image to form a binary segmentation mask.
The number of plant pixels are counted along the $x$ and $y$ axis in the binary segmentation mask to form two energy functions described below.
Since there are fewer plant pixels in the gaps between the plots, and the rows and ranges are relatively well aligned, the local minima of the energy functions correspond to the gaps between the plots and can be used to define the boundaries of the plots.

\begin{figure}[]
	\centering
	\centerline{\includegraphics[width = 0.4\textwidth]{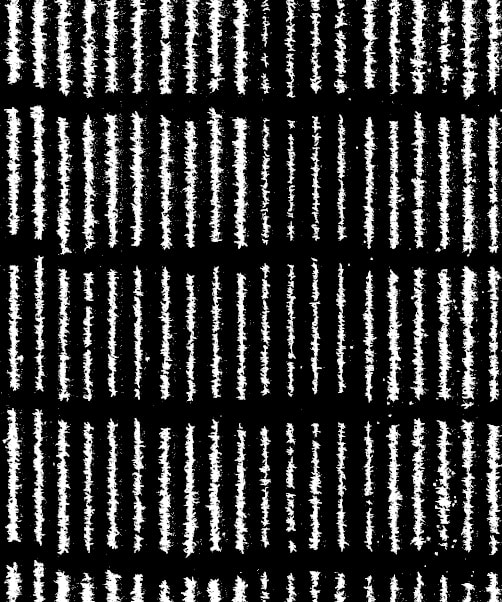}}
	\caption{A binary plant segmentation mask with approximately 5 ranges, 20 rows, and 100 plots}
\label{msk_img}
\end{figure}

\subsection{Energy Function and Range Separation}
The first step is to segment the plant pixels from the orthomosaic image $O(x,y,z)$ to form the plant segmentation mask $I(x,y)$.
A region of interest is extracted from the RGB orthomosaic image and converted to the HSV color space~\cite{javithesis,sharma2003}. 
The H color channel of the orthomosaic image is denoted as the image $O_H(x,y)$.
The binary plant segmentation mask $I(x,y)$ at $x$ and $y$ is estimated by 
thresholding the H channel image in an experimentally determined range of pixel values.
For the experiments described later, the pixel value range is $20$ to $90$: 

\begin{equation}
	\label{plt_height}
    I(x,y) = 
    \begin{cases}
    1 & {\text{if $20 \leq O_H(x,y) \leq 90$}} \\
     0 & \text{else}
    \end{cases}
\end{equation}
A sample plant segmentation mask is shown in Figure~\ref{msk_img}, note this image has approximately $5$ ranges, $20$ rows, and $100$ plots.
Note COPE is independent of the type of plant segmentation used.
Other plant segmentation methods such as Otsu~\cite{otsu_1979} can be used.
In addition, COPE is not limited to RGB orthomosaic images, any orthomosaic image  can be used.

We can then estimate the energy functions used for plot extraction.
From the plant segmentation mask $I(x,y)$, the range energy function $h_{ra}(y)$ at $y$ and global row energy function $h_{ro-gl}(x)$ at $x$ can be determined by counting the pixels in the plant segmentation mask along the $x$ and $y$ axis:

\begin{equation}
	\label{ran_prof}
    h_{ra}(y) = \sum_{x} I(x,y)
\end{equation}
\begin{equation}
	\label{row_prof}
    h_{ro-gl}(x) = \sum_{y} I(x,y)
\end{equation}


\begin{figure}[]
	\centering
	\centerline{\includegraphics[width = 0.43\textwidth]{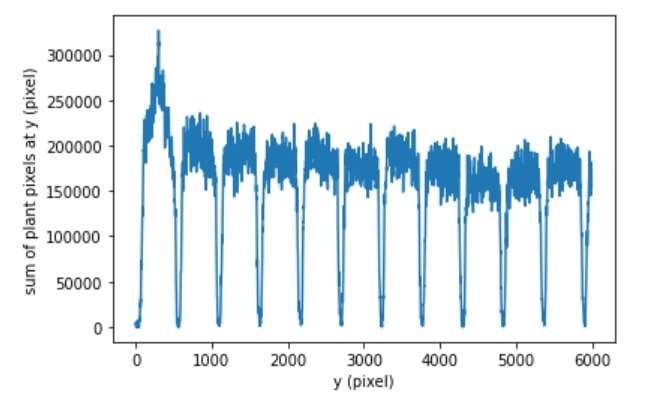}}
	\caption{A range energy function}
\label{example-range-profile}
\end{figure}

The energy functions can also be thought of as projections or profiles of the $x$ and $y$ spaces.
An example of a range energy function $h_{ra}(y)$ is shown in Figure~\ref{example-range-profile}.
Note the local minima correspond to the gaps between ranges.


\begin{figure}[]
	\centering
	\centerline{\includegraphics[width = 0.4\textwidth]{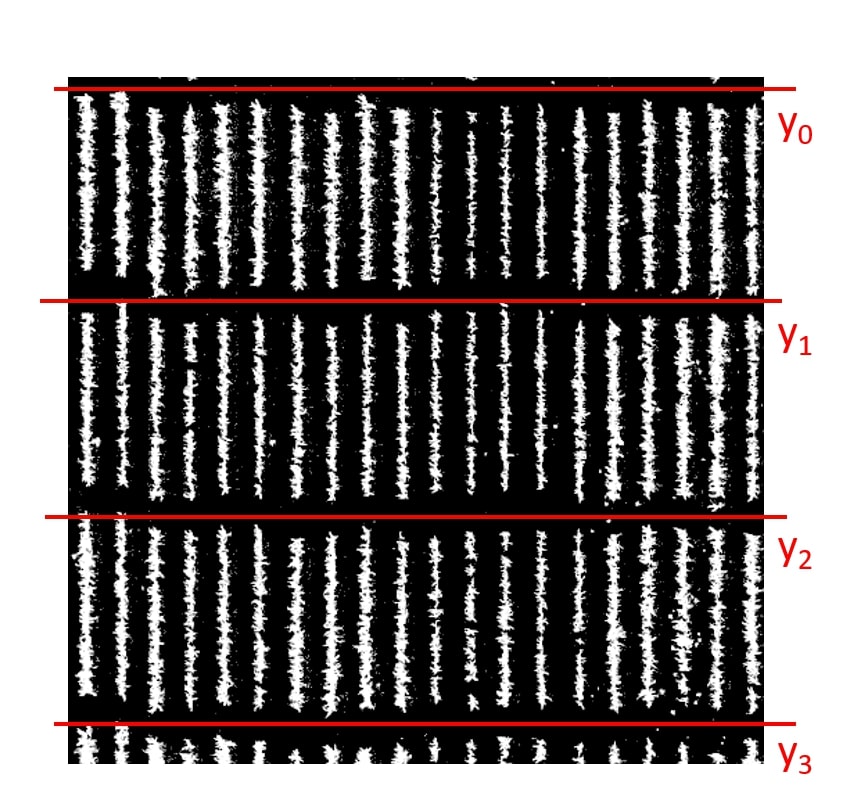}}
	\caption{A plant segmentation mask with range separation lines shown in red}
\label{ran_sep}
\end{figure}

Ranges are separated by assuming equal length and distance for all the ranges.
Assume the field is defined by $M$ rows $\times$ $N$ ranges.
The range separation lines can be defined as the set of $N+1$ horizontal lines that separate each range as shown in red in
Figure~\ref{ran_sep}.
We assume range separation lines to be $\Delta y$ apart.
We obtain the range separation lines from optimizing the range energy function $h_{ra}(y)$.
Therefore we need to find $\bar{y}_0$ and $\Delta y$.
Let the set of $N+1$ range separation lines be $\{ \bar{y}_0,  \bar{y}_1, \ldots , \bar{y}_{N} \}$, then

\begin{equation}
	\label{row_prof}
	\begin{aligned}
    & \bar{y}_n =\bar{y}_0 + n\bar{\Delta}y \\
     \bar{y}_0, \bar{\Delta}y = &\arg \min_{y_0,\Delta y} \sum \limits_{n = 0}^{N}
     h_{ra}(y_0+n\Delta y)
     \end{aligned}
\end{equation}

The equidistant assumption between the range separation lines we made above 
causes errors in the separation of the ranges due to the start/stopping of the seed drops when planting. 
Each range separation line is then adjusted to reduce the error.
The adjusted range separation lines can be defined as the set $ \{ \hat{y}_0,  \hat {y}_1, \ldots , \hat{y}_{N} \}$, then

\begin{equation}
	\label{row_prof}
	\begin{aligned}
     \hat{y}_i =\bar{y}_i + &\arg \min_{\Delta \bar{y}} h_{ra}(\bar{y}_i+\Delta \bar{y}),where\\
       &|\Delta \bar{y}| \leq D_{ran-gap}
     \end{aligned}
\end{equation}

$D_{ran-gap}$ is a constraint chosen to restrict the distance correction.
For our experiments  $D_{ran-gap} \approx 100$.


\subsection{Row Separation - Modified Comb Function}
With the ranges separated, the local row energy function $h_{ro}^{z}(x)$ of range $z$ at $x$ are obtained:
\begin{equation}
	\label{ran_prof}
    h_{ro}^{z}(x) = \sum\limits_{y = \hat{y}_{i-1}}^{\hat{y}_{i}} I(x,y) 
\end{equation}
Each range has its own local row energy function.

In order to smooth the optimization process, both the local and global row energy functions are normalized 0 to 1.
This results in a modified global row energy function $\hat{h}_{ro-gl}(x)$ and modified local row energy function $\hat{h}_{ro}^{i}(x)$\:

\begin{equation}
	\label{gl_prof}
    \hat{h}_{ro-gl}(x)= 
    \begin{cases}
    1 & {\text{if $h_{ro-gl}(x) \geq K$}} \\
    & \text{where K is the mean of $h_{rogl}(x)$}\\
    \frac{h_{rogl}(x)}{K} & \text{else}
    \end{cases}
\end{equation}

\begin{equation}
	\label{plt_height}
    \hat{h}_{ro}^{z}(x)= 
    \begin{cases}
    1 & {\text{if $h_{ro}^{i}(x) \geq K$}} \\
     & \text{where K is the mean of $h_{ro}^{z}(x)$}\\
    \frac{h_{ro}^{z}(x)}{K}& \text{else}
    \end{cases}
\end{equation}

\begin{figure}[]
	\centering
	\centerline{\includegraphics[width = 0.4\textwidth]{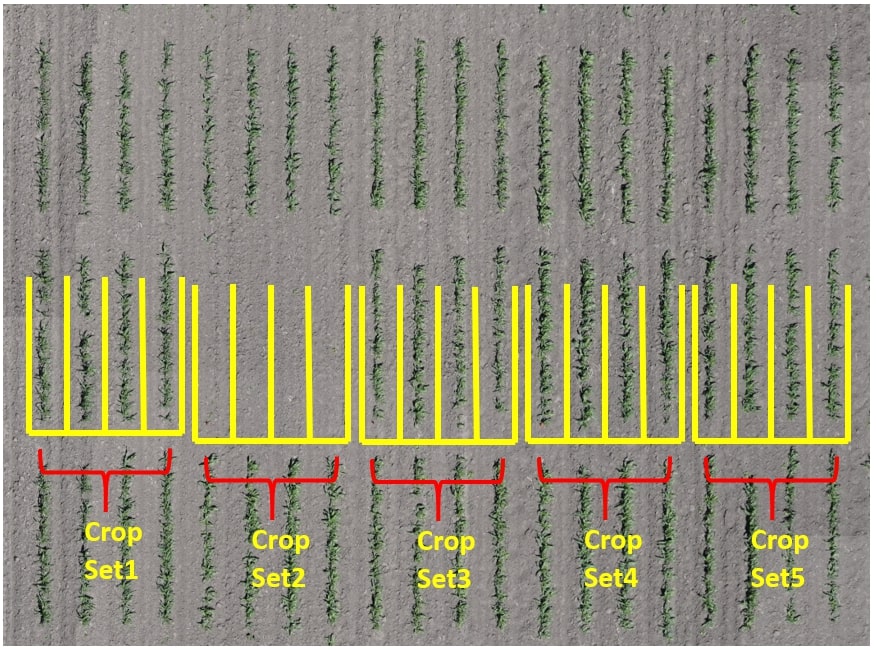}}
	\caption{Sets of crops (``crop sets'') relative to their range - note the comb-like structure overlaid on the image and shown in yellow}
\label{crop_set}
\end{figure}

\begin{figure}[]
	\centering
	\centerline{\includegraphics[width = 0.4\textwidth]{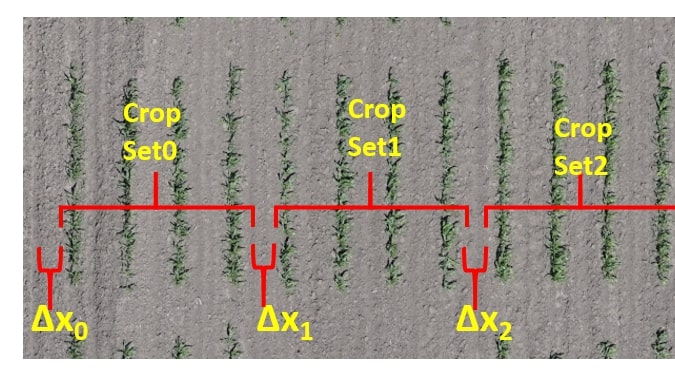}}
	\caption{Crop sets offsets relative to their range}
\label{crop_off}
\end{figure}

Recall from the previous section, the planter plants $C$ rows of crops at fixed distances apart simultaneously.
In each range, we refer to these $C$ rows  planted together as a ``crop set.''
An example of crop sets are shown in Figure~\ref{crop_set}.
Recall that $M$ is total number of rows in the range.
Each range can be represented by joining $\frac{M}{C}$ crop sets together in the range direction as shown in Figure~\ref{crop_set}. 
We refer to the distances between each crop set as the ``crop set offset'' denoted as $\Delta x_i$ where
$0 \leq i < \frac{M}{C}$. Sample crop set offsets are shown in Figure~\ref{crop_off}.
Note the positions of the plot boundaries inside any crop set are identical across all crop sets.
Even though the planter is GPS guided, 
the distances between the groups of crop sets vary by a small amount relative to the fixed distance between the individual plots inside the crop set provided by the planter.
In order to find the plot boundaries for the range, the only variable required is the crop set offsets.

\begin{figure}[]
	\centering
	\centerline{\includegraphics[width = 0.4\textwidth]{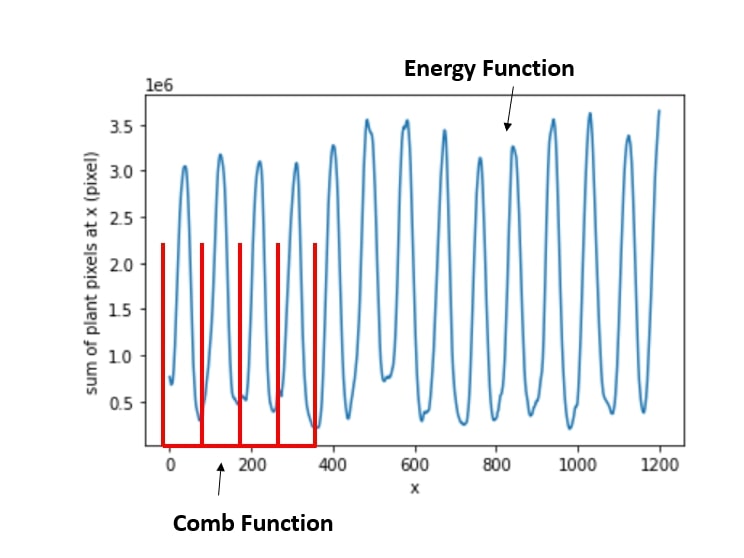}}
	\caption{The comb function spikes are fit to the local minima of the energy function}
\label{comb-on-energy}
\end{figure}

We need to locate the crop sets by using the energy functions to find the crop set offsets. 
Since a crop set contains plots with fixed boundaries, the corresponding local minima of the energy functions are a fixed distance apart. 
In order to find the local minima on the energy functions, a comb function $f(m)$ is formed. 
The comb function is defined by its width and the number of ``comb spikes.''
Examples of comb functions are shown in yellow in Figure~\ref{crop_set} where the combs are 300 pixels wide and there are 5 comb spikes.
Let the width of $f(m)$ be the same as the width of a crop set 
and the number of comb spikes to be $C+1$ with the comb spikes placed in the middle of the gaps between the 
plots (Figure~\ref{crop_set}).
The parameters of the comb function depends on the shape and dimension of the planting pattern as well as the orthomosaic image resolution.
We align the comb function with the energy function such that the comb spikes are at the local minimum as 
shown in Figure~\ref{comb-on-energy}.
The location of the crop set in the field can then be determined.

\begin{figure}[]
    \centering
    \subfloat[]{\label{Figure:input2}{\includegraphics[width = 0.4\textwidth]{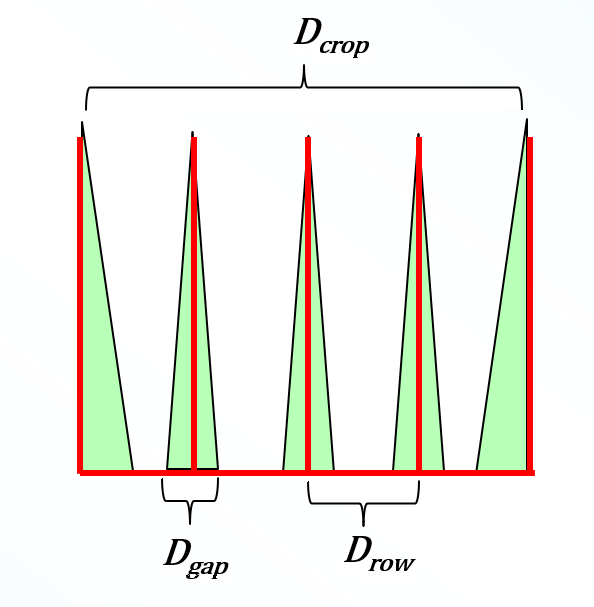}}}
    
    \subfloat[]{\label{Figure:prob2}{\includegraphics[width = 0.4\textwidth]{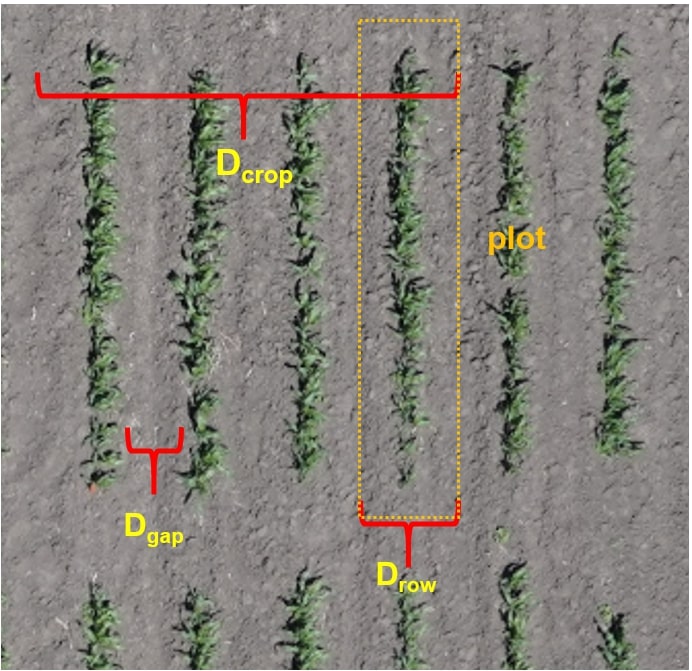}}}
       
    \caption{(a) Modified comb function with parameters used in Equation~\ref{deltax} (b) Modified comb function parameters overlaid on the field image}
    \label{c_func_param}
\end{figure}



Let $D_{gap}$ be the average gap (in pixels) width between plots in the same crop set as shown in Figure~\ref{c_func_param}.
$D_{gap}$ is used to modify the comb spikes to restrain the comb spikes to the gap between the plots.
The comb function $f(m)$ is altered to form the ``modified comb function'' $\hat{f}(m)$ by convolving the comb function with a triangle function to widen the comb spikes. 
The width of the triangle function is the same as the plot gap width $D_{gap}$ and the height is 1. 
The function is then trimmed to the same width as the crop set to form $\hat{f}(m)$ as shown Figure~\ref{c_func_param}.


Each range can be represented by joining $\frac{M}{C}$ crop sets together side by side,
with each crop set having a crop set offset of $\Delta x_i$ as shown in Figure~\ref{crop_off}.
To find the crop set offsets, starting at range $z$, we estimate first the crop set offset $\Delta x_0$, where:

\begin{equation}
	\label{deltax}
	\begin{aligned}
    \Delta x_i^z =arg&\min \limits_{\Delta x_i^z}{\omega_0\frac{{\Delta x_i^z}^2}{D_{row}^2}+\omega_1loss_1+\omega_2loss_2}\\
    \text{where }    &loss_1 = \frac{2}{D_{gap}}\hat{f}(n) \cdot \hat{h}_{ro}^{z}(n) \\
     &loss_2 = \frac{2}{D_{gap}}\hat{f}(n) \cdot \hat{h}_{ro-gl}(n) \\
     &|\Delta x_i| \leq D_{gap} \\
     & x_{off}^{i} + \Delta x_i \leq n \leq x_{off}^{i} + \Delta x_i + D_{crop}\\
     &\text{$x_{off}^{i} = 0$ when $i = 0$ }\\
     &\text{$x_{off}^{i}$ is an intermediate parameter } \\
     \end{aligned}
\end{equation} 

Where $D_{row}$ is the plot width, $D_{gap}$ is the plot gap width, $D_{crop}$ is the crop set width as shown in Figure~\ref{c_func_param}.
We then update $x_{off}^{i}$ by:

\begin{equation}
	\label{x_off}
	\begin{aligned}
      x_{off}^{i+1} = x_{off}^{i}+  \Delta x_i + D_{crop}\\
     \end{aligned}
\end{equation}

After finding $\Delta x_0$, we can find the rest of crop set offsets $\Delta x_i$ for range $z$ by a similar process.


\begin{figure}[]
	\centering
	\centerline{\includegraphics[width = 0.4\textwidth]{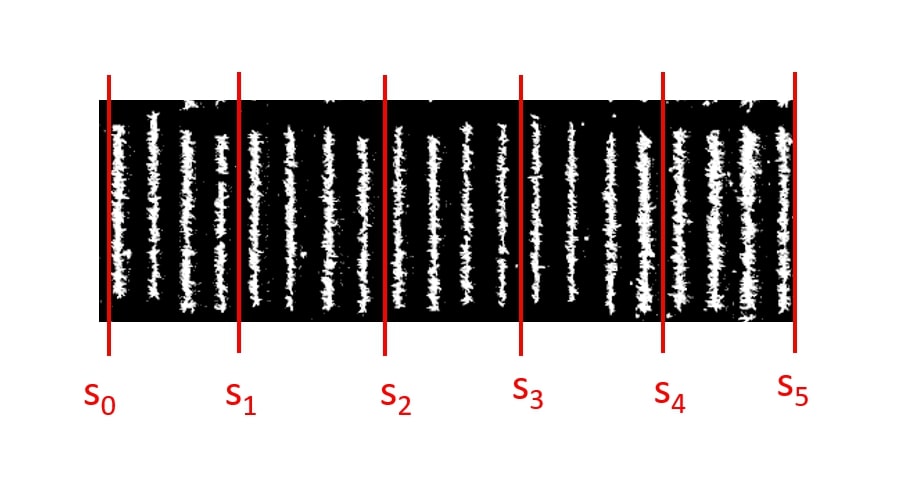}}
	\caption{A plant segmentation mask with crop set boundary lines in red}
\label{crop_sepe}
\end{figure}

\begin{figure}[]
	\centering
	\centerline{\includegraphics[width = 0.4\textwidth]{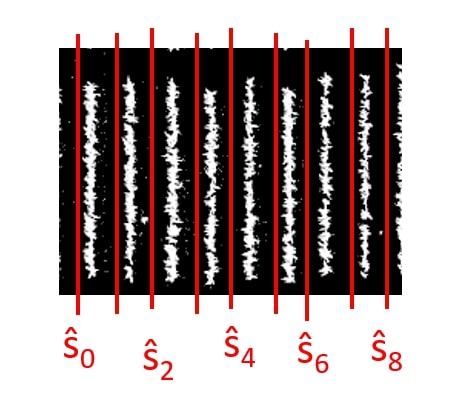}}
	\caption{A plant segmentation mask with plot boundary lines in red}
\label{plot_sepe}
\end{figure}

Crop set boundary lines are defined as the $\frac{M}{C}+1$ vertical lines that separate the crop sets from each other as shown in Figure~\ref{crop_sepe}. 
Let the set of crop set boundary lines be $\{ s_0, s_1, \ldots, s_{\frac{M}{C}}\}$, then:

\begin{equation}
	\label{crop_bound}
	s_j = 
	\begin{cases}
    \Delta x_0 & \text{if $j = 0$} \\
    x_{off}^{j-1}+\Delta x_{j-1}+D_{crop} & \text{if $j = \frac{M}{C}$} \\
    \frac{x_{off}^{j}+\Delta x_{j}+x_{off}^{j}}{2} & \text{else} \\
	\end{cases}
\end{equation}

The plot boundary lines are defined as the $M+1$ vertical lines that separate the plots from each other as shown in Figure~\ref{plot_sepe}.
Let the set of plot boundary lines be $ \{ \hat{s}_0, \hat{s}_1, \hat{s}_2, \ldots,  \hat{s}_M \}$, then
\begin{equation}
	\label{row_bound}
	\hat{s}_m = 
	\begin{cases}
    s_m & \text{if } m\mod C = 0 \\
    s_{m-k} + k D_{row}  & \text{else}\\
      \text{ where } k = m\mod C &  \\
	\end{cases}
\end{equation}
Using the plot boundary lines and the range separation lines, we can find plot boundaries for any plot in range $z$. 
For example if we examine the $i_{th}$ plot in range $z$: the row boundaries are $\hat{s}_{i-1}$ and $\hat{s}_{i}$, the range boundaries are  $\hat{y}_{z-1}$ and $\hat{y}_{z}$.
Repeat the steps above to find the plot boundaries for all the plots across all ranges.

\subsection{Boundary Fine-Tuning}
At this point all the plots from the same range have the same range boundaries.
The final step is to fine tune the range boundaries for each individual plot.
Assume a specific plot ($i_{th}$ row and $j_{th}$ range) with range boundaries $y_{top}^{ij}, y_{bot}^{ij}$ and row boundaries $ x_{left}^{ij}, x_{right}^{ij}$, the local range energy function $h_{local}^{ij}(y)$ at $y$ is then obtained:
\begin{equation}
	\label{crop_bound}
	h_{local}^{ij}(y) = 
	\begin{cases}
    \sum_{x} I(x,y) & \text{if $ x_{left}^{ij} \leq x \leq x_{right}^{ij}$ and} \\
     & \text{$y \leq y_{top}^{ij} -  D_{ran-gap}   $ and}  \\
     & \text{$y_{bot}^{ij} + D_{ran-gap} \leq y $ }  \\
      0 & \text{else} \\
	\end{cases}
\end{equation}

Recall $D_{ran-gap}$ is a constraint chosen to restrict the distance correction.
For our experiments  $D_{ran-gap} \approx 100$.
The adjusted local range energy function $\hat{s}_{ij}(y)$ is normalized  
using Equation~\ref{gl_prof}.
Similar to the modified comb function, a triangle function $t(y)$ with width of $D_{ran-gap}$ and height of $1$ is used to assist the optimization (Equation~\ref{bound_fine}).
Using the adjusted local range energy function, the fine-tuned $y_{topnew}^{ij}$ can be estimated  by
\begin{equation}
	\label{bound_fine}
	\begin{aligned}
     y_{topnew}^{ij} =y_{top}^{ij} + \arg \min_{\Delta y} & (\hat{h}_{local}^{ij} * t) (y_{top}^{ij}+\Delta y) \\ \text{where}
       &|\Delta y| \leq D_{ran-gap}\\
     \end{aligned}
\end{equation}
We use Equation \ref{bound_fine} to fine tune the range boundaries for all plots.

\begin{figure}[t]
	\centering
	\centerline{\includegraphics[width = 0.4\textwidth]{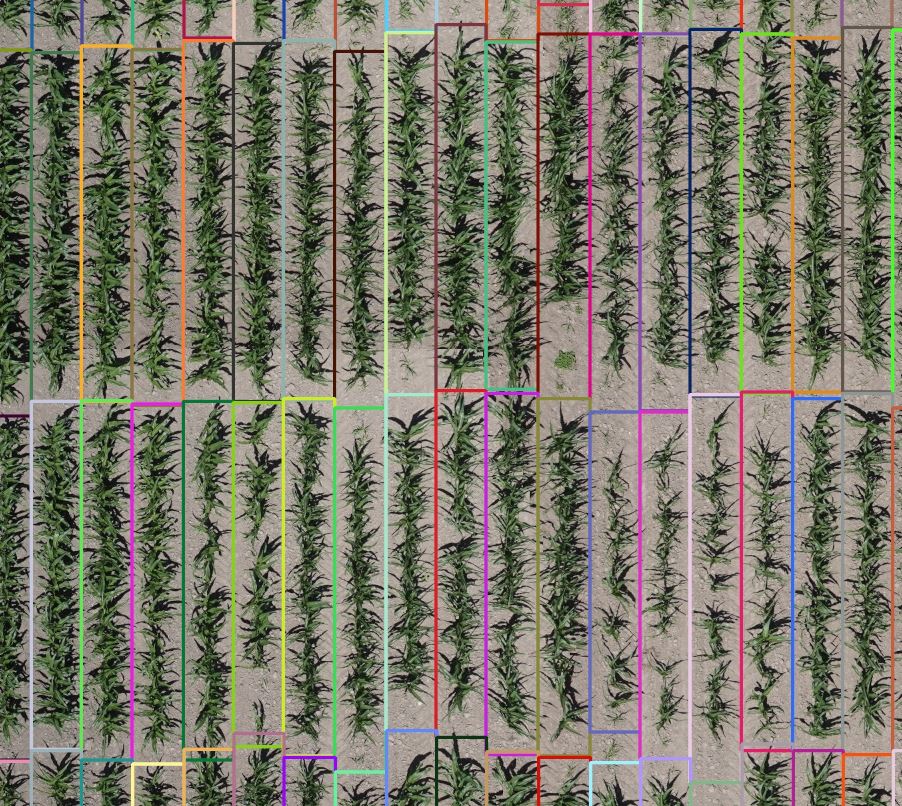}}
	\caption{Plots extracted by COPE from a maize orthomosaic image with 0.25cm/pixel resolution. The plots are color coded to make them easier to see}
\label{our_meth_result}
\end{figure}

\subsection{Summary of COPE}
COPE assumes the field is planted in a grid fashion. Additionally, COPE works better at the early stage of plant growth when the gap between the plants are not closed by the plants.
The requirements for COPE are:
\begin{itemize}
  \item Orthomosaic image
  \item Plant segmentation mask
  \item Region of interest from the orthomosaic image
  \item Number of row $M$ and ranges $N$ in the region of interest
  \item Range gap width $D_{ran-gap}$
  \item $C$ - the number of rows planted at the same time by the planter 
  \item Crop set width $D_{crop}$
  \item Plot width $D_{row}$
  \item Plot gap width $D_{gap}$
\end{itemize}
Note $C$, $D_{crop}$, $D_{row}$, and $D_{gap}$ are parameters used to describe the modified comb function. 
We only need to describe the modified comb function once unless the planting pattern or image resolution changes.
COPE will find the extracted plots, an example is shown in Figure~\ref{our_meth_result}.

\begin{figure}[]
    \centering
    \subfloat[]{\label{Figure:input2}{\includegraphics[width = 0.22\textwidth]{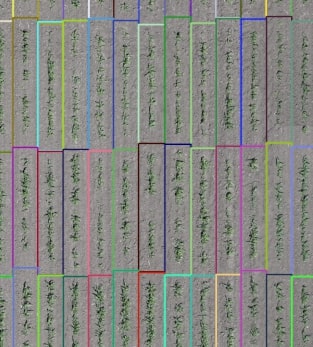}}}
    \quad    
    \subfloat[]{\label{Figure:input2}{\includegraphics[width = 0.22\textwidth]{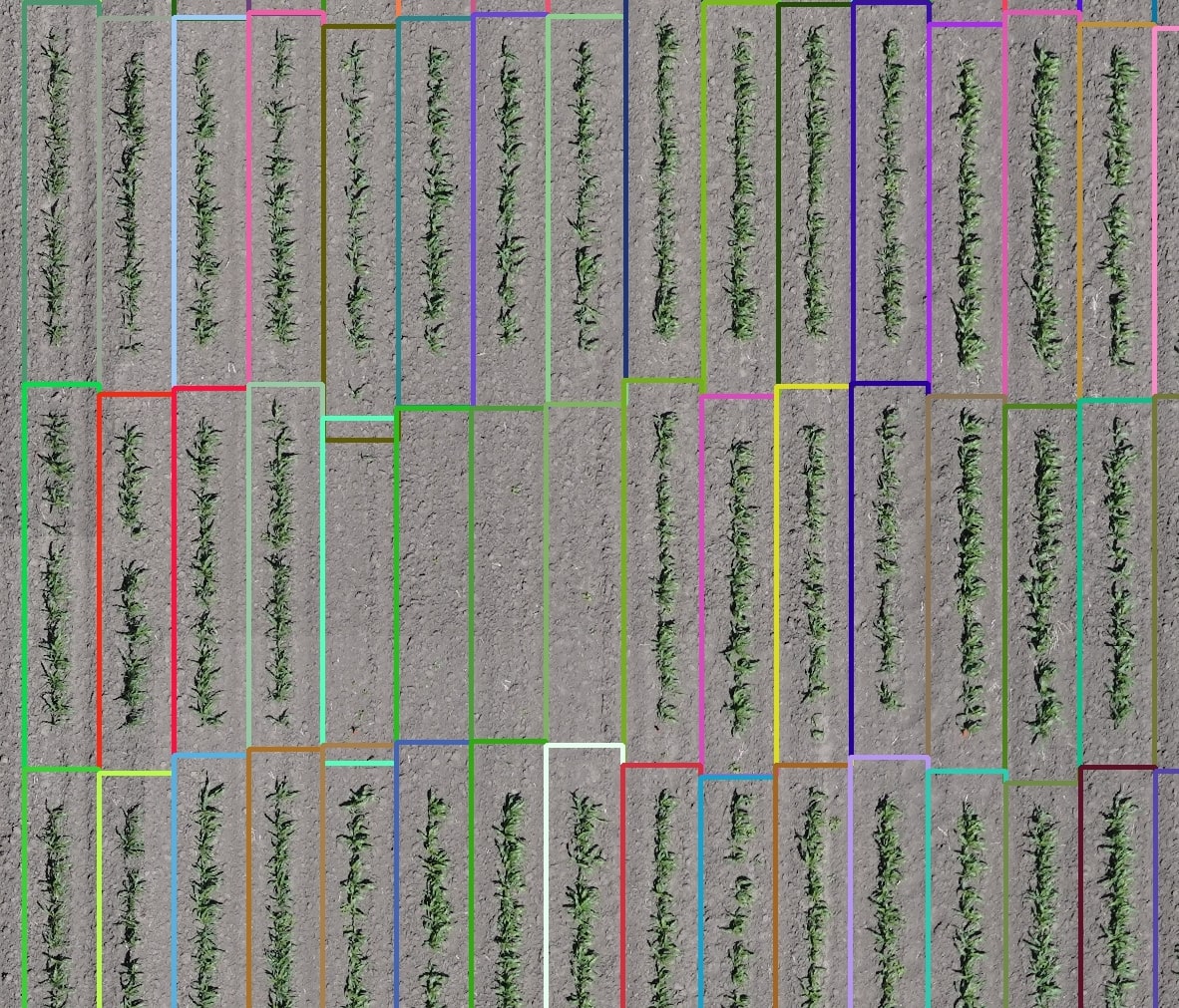}}}
    \quad    
    \subfloat[]{\label{Figure:prob2}{\includegraphics[width = 0.22\textwidth]{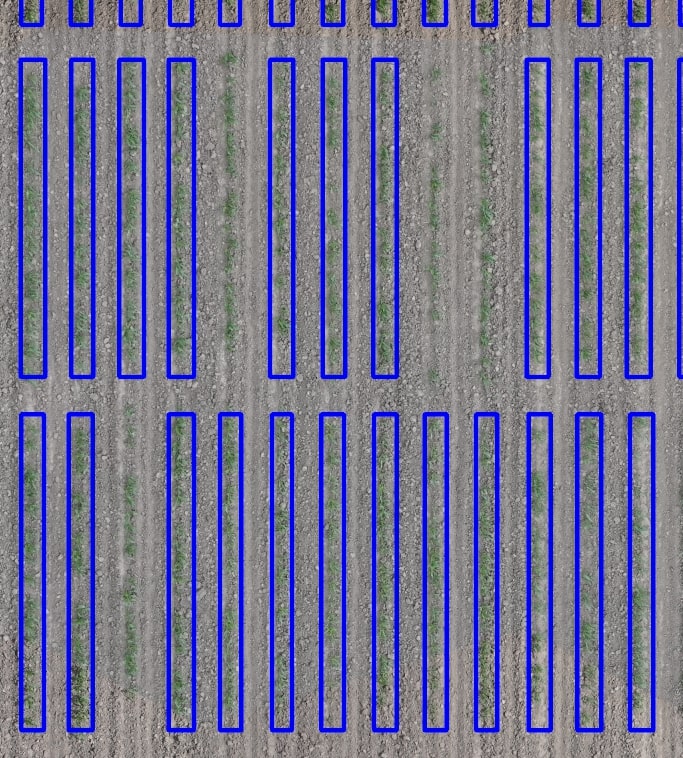}}}
    \quad 
    \subfloat[]{\label{Figure:prob2}{\includegraphics[width = 0.22\textwidth]{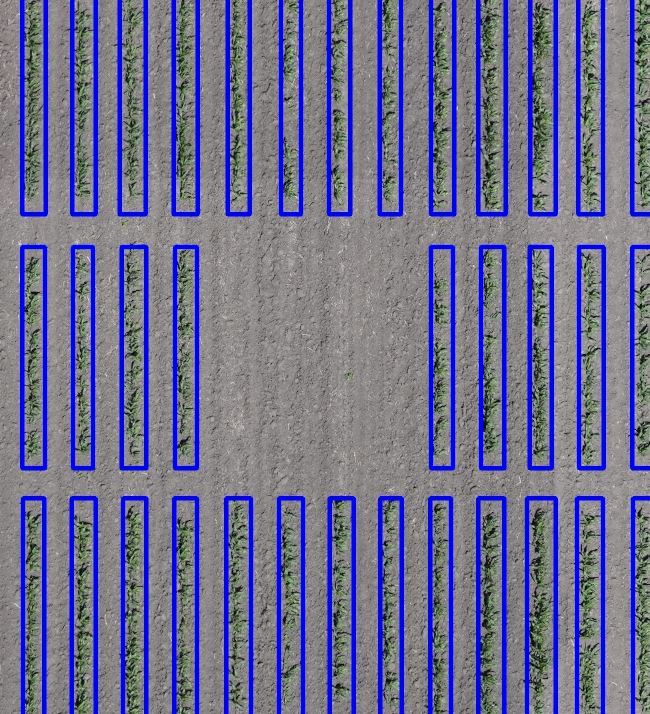}}}
    \quad 
    \subfloat[]{\label{Figure:prob2}{\includegraphics[width = 0.22\textwidth]{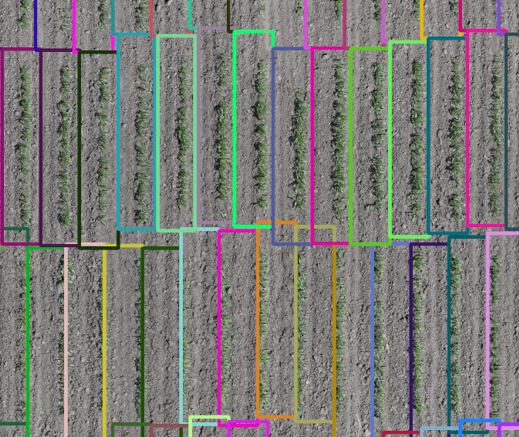}}}
    \quad 
    \subfloat[]{\label{Figure:prob2}{\includegraphics[width = 0.22\textwidth]{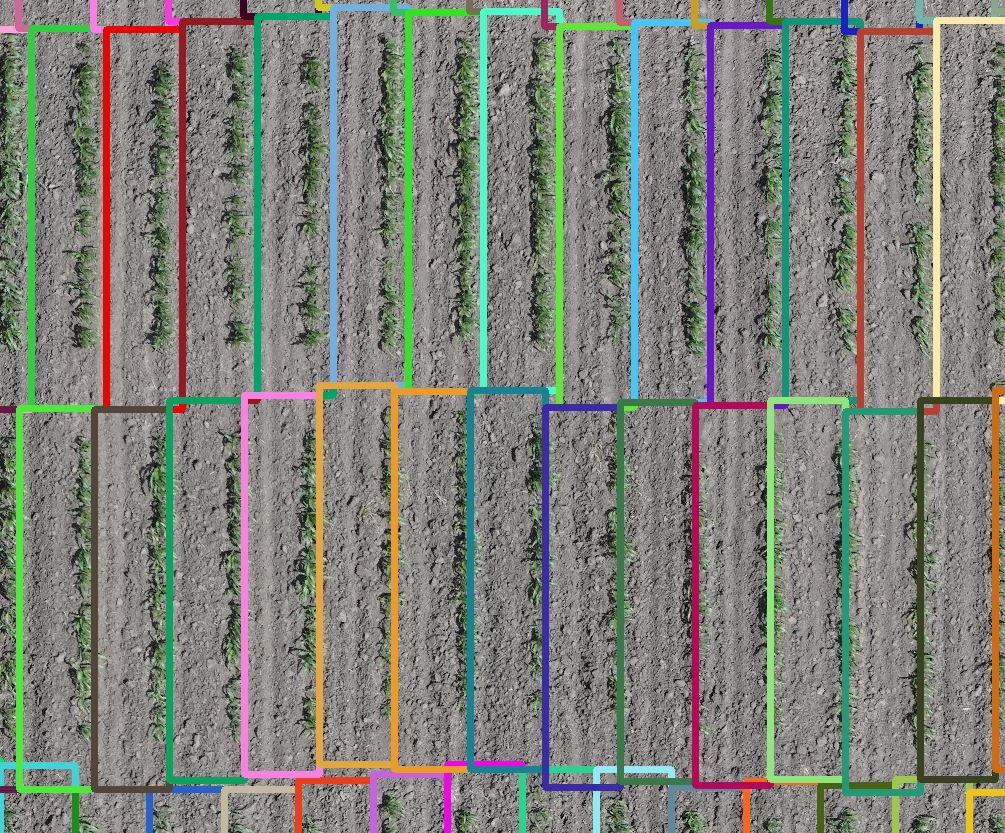}}}
    \quad 
    \caption{A comparison of plot extraction methods (sorghum image with 1cm/pixel spatial resolution)
             (a/b) COPE
             (c/d) EasyMPE~\cite{tresch2019}
             (e/f) The method in~\cite{javithesis}
}
    \label{Result_Comparison}
\end{figure}

\section{Experimental Results}
%

%

For our experiments we acquired orthomosaic images of sorghum and maize fields at various dates and altitudes.
We evaluated our method (COPE) against EasyMPE~\cite{tresch2019} and the method described in \cite{javithesis} using our dataset.
Note the typical orthomosaic images used in our experiment have more than 300 million pixels (e.g., $11500 \times 47500$ pixels).
Enlarged regions of an orthomosaic image are shown in Figure~\ref{Result_Comparison}.
From the orthomosaic images, five sub regions (approximately 700 plots) were randomly selected for manual labeling.
The manually labeled plot boundaries are used as ground truth for the evaluation of the three plot extraction methods.

The Intersection over Union (IoU)~\cite{powers_2011} is used to compare the extracted plots. 
For each of the 3 methods, we estimate the IoU between the extracted plots and the corresponding ground truth.

\begin{table}[h]
\centering
\begin{tabular}{lclll}
\toprule
Methods &  IoU\\
\midrule 
EasyMPE~\cite{tresch2019} & 0.48\\
Method in~\cite{javithesis} & 0.65 \\
COPE  &  \textbf{0.92} \\
\bottomrule
\end{tabular}
\caption{Average IoU for different plot extraction methods}
\label{f1_p_mate}
\end{table}

The average IoU for each method is shown in Table~\ref{f1_p_mate}. 
Our proposed method, COPE, has the highest IoU among all the methods.
From Figure~\ref{Result_Comparison}, it is also visually noted that COPE performs well. 
EasyMPE~\cite{tresch2019} has many missing plots and the method in \cite{javithesis} has plot boundaries intersecting with the plant material.

\begin{table}[h]
\centering
\begin{tabular}{lclll}
\toprule
Methods & PBA Time \\
\midrule 
EasyMPE~\cite{tresch2019} & 1659s\\
Method in~\cite{javithesis} & 3712s\\
COPE  &  \textbf{438s} \\
\bottomrule
\end{tabular}
\caption{Manual plot boundary adjustment (PBA) time in seconds(s) for 700 plots}
\label{Methods_time}
\end{table}

Manual plot boundary adjustment is sometimes needed to correct boundary errors of the plot segments in the 
extracted plots~\cite{yang2021integrating}. 
An experienced  post-doctoral researcher, not associated with our team, evaluated each method by estimating the total time required for manual adjustment. 
For the five sub regions used for ground truthing (700 plots) the plot boundaries found for each of the three methods
were manually adjusted by the researcher.
The manual plot boundary adjustment (PBA) time for each method for the 700 plots is shown in Table~\ref{Methods_time}.
COPE achieves more than a $70\%$ reduction in plot boundary adjustment time compared to the 
second best method EasyMPE. 
Note lower manual plot boundary adjustment time indicates many of the extracted plots need no adjustment which corresponds to
better automated extraction of the plots.

\section{Conclusions and Future Work}
Overall COPE is a reliable tool for plot extraction using RGB orthomosaic images.
COPE achieves the highest performance in terms of IoU while also requiring the least amount of manual adjustment time. 
COPE currently requires fields to be planted in a strictly grid-like pattern.  
In the future, we will improve our method to reduce the reliance on strictly grid-like fields, for example when the the grids are not well aligned in the field. We will also address imagery other than RGB images as well as imagery with different lighting conditions.


The source code for COPE and the datasets used for our experiments will made available to the community.
For more information contact Edward J. Delp at
ace@ecn.purdue.edu

\section{Acknowledgments}
We thank Professor Ayman Habib and the Digital Photogrammetry Research Group (DPRG) from the School of
Civil Engineering at Purdue University for providing the
orthomosaic images used in this paper. 
We also thank Jieqiong Zhao from Arizona State University for helping with the groundtruthing.
The work presented here was
funded by the Advanced Research Projects Agency-Energy (ARPA-E), U.S. Department of Energy, under Award
Number DE-AR0001135. The views and opinions of the
authors expressed herein do not necessarily state or reflect those of the United States Government or any agency
thereof. Address all correspondence to Edward J. Delp,
ace@ecn.purdue.edu

{\small
\bibliographystyle{ieee_fullname}
\bibliography{ref}

\begin{thebibliography}{10}\itemsep=-1pt

\bibitem{araus2013}
J. Araus and J. Cairns.
\newblock Field high-throughput phenotyping: The new crop breeding frontier.
\newblock {\em Trends in Plant Science}, 19, October 2013.

\bibitem{costa2019}
C. Costa, U. Schurr, F. Loreto, P. Menesatti, and S. Carpentier.
\newblock Plant phenotyping research trends, a science mapping approach.
\newblock {\em Frontiers in Plant Science}, 9:1933, 2019.

\bibitem{habib2016}
A. Habib, W. Xiong, F. He, H.~L. Yang, and M. Crawford.
\newblock Improving orthorectification of uav-based push-broom scanner imagery
  using derived orthophotos from frame cameras.
\newblock {\em IEEE Journal of Selected Topics in Applied Earth Observations
  and Remote Sensing}, 10:1--15, 02 2016.

\bibitem{johansen2019}
K. Johansen, M. Morton, Y. Malbeteau, B. Aragon, S. AlMashharawi, M. Ziliani,
  Y. Angel, G. Fiene, S. Negrao, M. Mousa, M. Tester, and M. McCabe.
\newblock Unmanned aerial vehicle-based phenotyping using morphometric and
  spectral analysis can quantify responses of wild tomato plants to salinity
  stress.
\newblock {\em Frontiers in Plant Science}, 10, 03 2019.

\bibitem{khan2019}
Z. Khan and S. Miklavcic.
\newblock An automatic field plot extraction method from aerial orthomosaic
  images.
\newblock {\em Frontiers in Plant Science}, 10, 05 2019.

\bibitem{li_2014}
L. Li, Q. Zhang, and D. Huang.
\newblock A review of imaging techniques for plant phenotyping.
\newblock {\em Sensors}, 14(11):20078--20111, 2014.

\bibitem{lin2021}
Y. Lin, T. Zhou, T. Wang, M. Crawford, and A. Habib.
\newblock New orthophoto generation strategies from uav and ground remote
  sensing platforms for high-throughput phenotyping.
\newblock {\em Remote Sensing}, 13:860, 02 2021.

\bibitem{otsu_1979}
N. Otsu.
\newblock A threshold selection method from gray-level histograms.
\newblock {\em IEEE Transactions on Systems, Man, and Cybernetics}, 9:62--66,
  Janurary 1979.

\bibitem{powers_2011}
D. Powers.
\newblock Evaluation: From precision, recall and f-factor to roc, informedness,
  markedness \& correlation.
\newblock {\em Journal of Machine Learning Technologies}, 2(1):37--63, 2011.

\bibitem{javithesis}
J.~R. Prat.
\newblock {\em Image-Based Plant Phenotyping Using Machine Learning}.
\newblock PhD thesis, Purdue University, West Lafayette, IN, May 2019.
\newblock Available at https://doi.org/10.25394/PGS.7774313.v1.

\bibitem{sharma2003}
Gaurav Sharma.
\newblock {\em Digital Color Imaging Handbook}.
\newblock CRC Press, Boca Raton, FL, 2003.

\bibitem{tang2021}
Z. Tang, A. Parajuli, C.~J. Chen, Y. Hu, S. Revolinski, C.~A. Medina, S. Lin,
  Z. Zhang, and L. Yu.
\newblock Validation of uav-based alfalfa biomass predictability using
  photogrammetry with fully automatic plot segmentation.
\newblock {\em Scientific Reports}, 11:3336, 02 2021.

\bibitem{tresch2019}
L. Tresch, Y. Mu, A. Itoh, A. Kaga, K. Taguchi, M. Hirafuji, S. Ninomiya, and
  W. Guo.
\newblock Easy mpe: Extraction of quality microplot images for uav-based
  high-throughput field phenotyping.
\newblock {\em Plant Phenomics}, 2019:1--9, 11 2019.

\bibitem{vanderlip_1972}
R.~L. Vanderlip.
\newblock {\em How a Sorghum Plant Develops}.
\newblock Kansas State University, Manhattan, KS, 1972.

\bibitem{wiesnerhanks2019}
T. Wiesner-Hanks, H. Wu, E. Stewart, C. DeChant, N. Kaczmar, H. Lipson, M.
  Gore, and R. Nelson.
\newblock Millimeter-level plant disease detection from aerial photographs via
  deep learning and crowdsourced data.
\newblock {\em Frontiers in Plant Science}, 10, 2019.

\bibitem{xie2020}
C. Xie and C. Yang.
\newblock A review on plant high-throughput phenotyping traits using uav-based
  sensors.
\newblock {\em Computers and Electronics in Agriculture}, 178:105731, 11 2020.

\bibitem{yang2021integrating}
K. Yang, S. Chapman, N. Carpenter, G. Hammer, G. McLean, B. Zheng, Y. Chen,
  E.J. Delp, A. Masjedi, M. Crawford, et~al.
\newblock Integrating crop growth models with remote sensing for predicting
  biomass yield of sorghum.
\newblock {\em in silico Plants}, 3(1):1, 2021.

\bibitem{yang2020high}
W. Yang, H. Feng, X. Zhang, J. Zhang, J. Doonan, W. Batchelor, L. Xiong, and J.
  Yan.
\newblock Crop phenomics and high-throughput phenotyping: Past decades, current
  challenges, and future perspectives.
\newblock {\em Molecular Plant}, 13, 01 2020.

\end{thebibliography}
}

\end{document}